\documentclass[journal]{IEEEtran}

\usepackage[utf8]{inputenc}
\usepackage[T1]{fontenc}
\usepackage{siunitx}
\usepackage[colorlinks=true, allcolors=blue]{hyperref}
\usepackage{amsmath}
\usepackage{amsfonts}
\usepackage{soul} 
\usepackage{caption}
\usepackage{subfigure}

\usepackage{tikz,xcolor}
\definecolor{lime}{HTML}{A6CE39}
\DeclareRobustCommand{\orcidicon}
{
    \begin{tikzpicture}
    \draw[lime, fill=lime] (0,0) circle [radius=0.16] 
    node[white] {{\fontfamily{qag}\selectfont \tiny ID}};    \draw[white, fill=white] (-0.0625,0.095) circle [radius=0.007];    
    \end{tikzpicture}
    \hspace{0mm}}
\foreach \x in {A, ..., Z}{%
    \expandafter\xdef\csname orcid\x\endcsname{\noexpand\href{https://orcid.org/\csname orcidauthor\x\endcsname}{\noexpand\orcidicon}}
}

\begin{document}

\title{Intelligent Perception System for Vehicle-Road Cooperation}

\author{Songbin Chen
}

\markboth{Journal of \LaTeX\ Class Files,~Vol.
}%
{Shell \MakeLowercase{\textit{et al.}}: Bare Demo of IEEEtran.cls for IEEE Journals}

\maketitle

\begin{abstract}
With the development of autonomous driving, the improvement of autonomous driving technology for individual vehicles has reached the bottleneck. The advancement of vehicle-road cooperation autonomous driving technology can expand the vehicle's perception range, supplement the perception blind area and improve the perception accuracy, to promote the development of autonomous driving technology and achieve vehicle-road integration. This project mainly uses lidar to develop data fusion schemes to realize the sharing and combination of vehicle and road equipment data and achieve the detection and tracking of dynamic targets. At the same time, some test scenarios for the vehicle-road cooperative system were designed and used to test our vehicle-road cooperative awareness system, which proved the advantages of vehicle-road cooperative autonomous driving over single-vehicle autonomous driving.
\end{abstract}


%
\IEEEpeerreviewmaketitle

\section{Introduction}



\subsection{Background}
Autonomous driving refers to a technology that replaces human driving operations by enabling cars to be equipped with advanced onboard sensors, controllers, actuators, and other devices to have complex environment perception, intelligent decision-making, and precise control. The use of automatic driving technology can not only avoid traffic accidents caused by driver error but also make up for the shortcomings of human drivers in the perception of environmental situations, decision-making driving behavior, and other aspects. It is an industry trend to use autonomous driving to solve problems such as traffic congestion and traffic accidents, which can greatly improve the convenience of travel and improve the efficiency of cities.

At present, the existing research on autonomous driving mainly focuses on the intelligence of a single-vehicle. Autonomous vehicles use their sensors, such as lidar and cameras, to sense the surrounding environment and make control decisions through their algorithms. Vehicle-road cooperation is a new research direction for autonomous driving. It can jointly perceive the road traffic environment in real-time through the sensors of the vehicle end and the road end, and exchange and share the perceived information to enhance the perception and prediction ability of autonomous vehicles.

In recent years, the perception ability and computing ability of autonomous driving of a single-vehicle have been developed very advanced, almost reaching the development bottleneck, but its limitations are still obvious, so the development of single-vehicle intelligence can not solve the problem of autonomous driving once and for all. The vehicle-road cooperation autonomous driving provides multi-field and multi-angle perception, which can not only expand the range of perception of the vehicle but also overcome some defects left by a single-vehicle. However, in the field of autonomous driving, there is still a big gap in the research on vehicle-road cooperation.

\subsection{Motivation}
The blind field of the visual field of a single-vehicle affects the safety of autonomous driving. Since the autonomous driving of a single-vehicle only senses from the perspective of the vehicle, it is likely to face the situation that the target object is blocked by other vehicles or buildings, resulting in the lack of perception of the target object, which is easy to cause traffic accidents. The vehicle-road cooperation can solve the problem of the blind area of vehicle vision well. By placing sensors in different places on the road, it can obtain different perception perspectives. By sharing data with vehicles, it can expand the perception range and reduce the blind area of vision, which can greatly reduce traffic accidents caused by insufficient perception and improve the safety of autonomous driving.

The potential sensor failure of autonomous driving affects the reliability of autonomous driving. Every sensor on the vehicle plays an important role. Once a sensor fails, it can lead to perceptual failure in a certain direction or even the whole system crash. Vehicle-road cooperation can solve the problem of vehicle sensor failure well. It can carry out multiple perceptions of an object through the multi-angle perception of road conditions, providing redundancy for the perception system. Even if the vehicle sensor fails, it can also sense the target object through road end equipment. This can greatly reduce perceptual failures and improve the reliability of autopilot systems.

\subsection{Objective}
We want to design and implement a vehicle-road cooperative system in terms of perception. At the same time, a series of tests are designed to test our perception system.

To realize information sharing, we need to realize the fusion of vehicle and road perception information to some extent. The data obtained by different sensors at the vehicle end and the road ends are distributed in different perspectives and features, which leads to that the data cannot be directly used by one end for perception or decision-making. This requires a mechanism to unify perception data into one domain.

To verify the superiority of our vehicle-road cooperative system, we will design a series of experimental scenes and evaluation indexes for the system. Existing autonomous driving scenarios cannot support the comprehensiveness and effectiveness of vehicle-road cooperation testing, and cannot correctly reflect the advantages and disadvantages of the vehicle-road cooperation systems. This requires a new series of test scenarios and evaluation indicators for vehicle-road cooperation.

\section{Related Work}
\label{sec:related_work}


Intelligent transportation systems (ITS) are systems that use cooperation technologies and systems engineering concepts to develop and improve various transportation systems. To build the network of ITS, four architectures have been proposed: vehicle-to-vehicle (V2V), vehicle-to-infrastructure (V2I), vehicle-to-pedestrian (V2P), and vehicle-to-Anything (V2X) \cite{WangLonghe2019Vcci}. V2V communication is implemented between moving vehicles, which act as source, destination, and router during communication. Intermediate nodes (vehicles) transfer messages between the source and destination nodes. V2I communication allows vehicles to communicate with roadside infrastructure \cite{V2I,lan2022semantic}. In this case, the vehicle acts as both source and destination. V2P allows direct, instant, and flexible communication between mobile vehicles and roadside passengers. Through portable wireless devices, passengers can easily join VANET as roadside nodes to express their travel needs \cite{lan2022class,V2P,lan2018directed}. V2X provides vehicle-to-network and vehicle-to-pedestrian communication services \cite{V2X}.

Wu X et al. proposed a new multimodal framework, SFD, to solve the problem of multimodal fusion between point clouds \cite{xu2019online} and images \cite{wu2022sparse}. Wu H et al. designed the first tracklet proposal network PC-TCNN, which significantly improved multi-object tracking performance by introducing tracklet proposal design \cite{wu2021tracklet}.
\cite{lan2019evolutionary,lan2019simulated} proposed the simulation platform for swarm robotics.
Xiao Z et al. significantly improved the integrity of target perception through the sharing of sensor information and the utilization of map data \cite{s19091967}. Wang Z et al. show different sensor fusion strategies and discuss the method of establishing motion model and data association in multi-target tracking \cite{8943388,lan2017development,lan2016developmentVR}. Vishnukumar H J et al. think that artificial intelligence could be used to generate simulated scenes and verify results in autonomous driving \cite{AITEST}.

\section{Methodology}
\label{sec:methodology}


\begin{figure*}[htb]
    \centering
    \includegraphics[width=.9\textwidth]{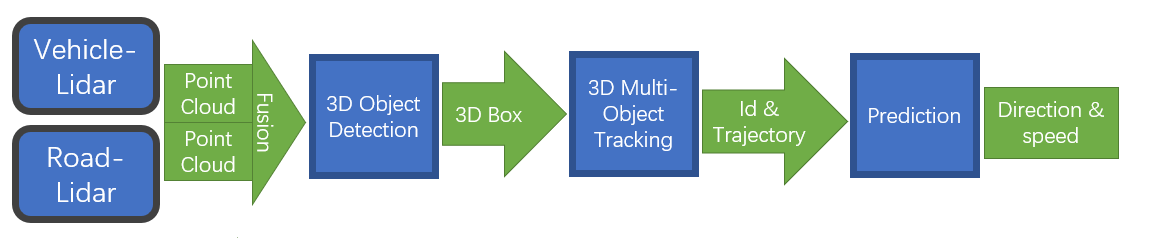}
    \caption{The main workflow of the perception system.}
\end{figure*}

\subsection{System Flow}

\subsubsection{Sensor Setting}
To perceive the environment, we need to set up sensors at the vehicle end and the road end to obtain environmental information. At the beginning of the whole process, the selection and setting of sensors have a decisive impact on the subsequent treatment and effect.

To determine the vehicle position, the most direct way is to obtain the longitude and latitude coordinates provided by the satellite positioning system, but in reality, the signal may be weak or drift due to environmental interference. Therefore, the vehicle position can be deduced by using a satellite positioning sensor, inertial measurement unit, and other data comprehensively to improve the accuracy and robustness of vehicle position implementation.

To observe the external environment around the vehicle, various lidar sensors, vision sensors, and V2X communication equipment are arranged in the rear of the vehicle to observe meteorological conditions, and road conditions and identify traffic signs, as well as obstacles and traffic participants identification, tracking and prediction.

Considering the vehicle-road cooperation system, in the vehicle end, usually adopt lidar as its main environmental sensors, install it on or around the roof, because laser radar usually has a 360 - degree Angle and have depth information, within a few hundred meters can simultaneously detect the contour of the object, at the same time with the reflectivity of the material of information can also be used to identify objects. However, the disadvantage of lidar is that it has the problem of insufficient resolution for distant objects and insufficient precision for close objects. A camera is also used as a sensor on the car end, which is installed in the front of the vehicle because the camera can obtain visible light information reflected by objects in a certain direction. The disadvantage is that ordinary cameras lack depth information and are easily confused by colors.

Lidar or camera sensors are also used at the road end, but they are also different in many ways. Road end sensors are installed at a higher height than vehicle end sensors for better visibility and are usually installed on poles at intersections and roadsides. Because of its particularity, the road end sensor also needs a longer sensing range and precision than the car end sensor. Moreover, the device at the end of the road will not move after installation, so the data characteristics obtained by the sensor are quite different from those of the vehicle in motion.

\subsubsection{Object Detection}
After obtaining the environmental information obtained by the sensor, to remove the unimportant part of the environmental information and extract useful information, we need to detect objects in the data, such as pedestrians, vehicles, lane lines, and so on.

For pedestrian and vehicle recognition, the target detection model is used to detect point cloud data obtained by lidar. For the recognition of point cloud, one approach regards point cloud as a sequence according to the scanning order and makes use of the continuity of time when the object is scanned, which can be recognized by R-CNN. Another approach is to mark point clouds as voxels in space, which can be identified by Voxel-CNN by taking advantage of the spatial proximity of objects. In addition, the image data obtained by the camera can also be detected by the detection model. Ultimately we want to get boundary information between pedestrians and vehicles.

For man-made information such as lane lines, road signs, and traffic lights, processing image data can be easier to detect than processing point cloud data. The mathematical characteristics of lane lines are obvious, so the track of lane lines can be extracted by the Hough transform, or the distribution of lane lines in space can be directly extracted by the neural network model. For road signs and traffic lights, convolutional neural networks can be used for specific recognition \cite{lan2019evolving,lan2018real}.

In addition, relying on the advantage of the position of the sensor at the road end not changing, static elimination of data can also be tried, which can remove the motionless objects and leave the information of the moving objects that we care about, which can better extract the information of the moving objects.

\subsubsection{Object Tracking and Prediction}
After object detection, to distinguish the threat degree of vehicles, achieve better avoidance and improve driving efficiency, we need to restore the target's actions and deduce the target's motion state at the next moment. Therefore, we trace the detected objects, trace the information of each object at each point in time, and get the trajectory of the object.

For objects that can move, such as pedestrians and vehicles, after they are detected from the original data, they can be simply processed by a filtering algorithm. The object information is input into the algorithm in chronological order, and the algorithm will judge whether a new object belongs to the same as the previous one according to the position and speed of each object at the previous moment, and can assign the same ID to the same object \cite{lan2022time,lan2021learning2,lan2021learning1}. The same ID indicates that these objects are the same object at different moments. By placing the positions of objects with the same ID in chronological order, the trajectory of the object can be obtained.

Then we need to predict the future motion according to the trajectory, mainly including the direction and speed of motion. This can also often be done by filtering algorithms. New estimates can be generated from historical states combined with current states.

\subsection{Data Fusion}
After designing the perception process, we need to design the scheme of vehicle-road collaboration, among which the most important way of collaboration is the sharing and fusion of vehicle-road perception data. We design two data fusion schemes.

\subsubsection{Pixel-level Fusion}
Pixel-level fusion is to directly share and combine the original data of sensors at the vehicle end and the road end and map each data point to the same space.

For point cloud data obtained by lidar, each data point has 3D coordinate information. To carry out a pixel-level fusion of point cloud data of road end and vehicle end radar, what we need to do is to unify the coordinate system of logarithm points.

The data fused at pixel level needs special processing methods. Different from the lidar point cloud of a single vehicle, the fused point cloud has two sources, and the usual object detection methods cannot make good use of the information in it \cite{xu2019online}. Therefore, to exploit the advantages of pixel-level fusion, a new object detection model needs to be proposed.

Pixel-level fusion has the advantage of high precision. However, the disadvantage is that the amount of data to be processed is large and it is difficult to fuse the data of different types of sensors.

\subsubsection{Feature-level Fusion}
Feature-level fusion is to share and combine the extracted features by processing sensor data separately.

For the 3D boundary of pedestrians and vehicles identified, each boundary consists of eight coordinate points. To carry out a feature-level fusion of the 3D boundary of pedestrian and vehicle recognized by road end and vehicle end, what we need to do is to unify the coordinate system of the boundary. Then, the boundary of the same kind of object with a high overlap degree will be considered as the same object and unified into one.

The feature-level fusion data can be easily utilized in the existing autonomous driving framework. After the fusion of the boundary obtained from 3D object detection of point cloud data, the quantity and quality of the perceived boundary can be improved, and the subsequent processing method will not change much. Therefore, feature-level fusion should be easily used by autonomous driving systems.

The advantage of feature-level fusion is that it is convenient to implement, and the same feature extracted from different sensors can also be fused. However, the disadvantage is that some information will be lost after feature extraction.

\subsection{Implementation}

To verify the feasibility of our vehicle-road cooperation intelligent perception system, we made a simple implementation for each part of the design. 

CARLA is an open-source simulator for autonomous driving research that supports the development, training, and validation of autonomous driving systems\cite{CARLA}. Considering cost and security, we choose the CARLA simulation platform to build our system.

\subsubsection{Sensor Setting}

\begin{enumerate}
    \item We chose lidar as the primary sensor on the vehicle to sense pedestrians and vehicles, which was set to a sensing range of 20 meters and mounted on top of the car 2 meters above the ground.
    \item Lidar is also used at the end of the road, which is installed 3 meters above the intersection and has a wider sensing range of up to 40 meters.
    \item Two GNSS modules are distributed in the front and rear of the vehicle so that the vehicle rotation Angle can be calculated while obtaining the vehicle position.
    \begin{equation}
        (x, y) = (x_1 / 2, y_1 / 2) + (x_2 / 2, y_2 / 2)
    \end{equation}
    \begin{equation}
        yaw = \arctan((y_2-y_1)/(x_2-x_1))
    \end{equation}
    \item Additional cameras are mounted on the roof, looking forward, to extract lane line information.
\end{enumerate}

\subsubsection{Object Detection}

\begin{figure}[!ht]
    \centering
    \subfigure[Vehicle and pedestrian detection results of the point cloud using point-RCNN. White dots are point clouds, and green, blue, and yellow boxes respectively represent vehicles, bicycles, and pedestrians.]{
        \includegraphics[width=.45\textwidth]{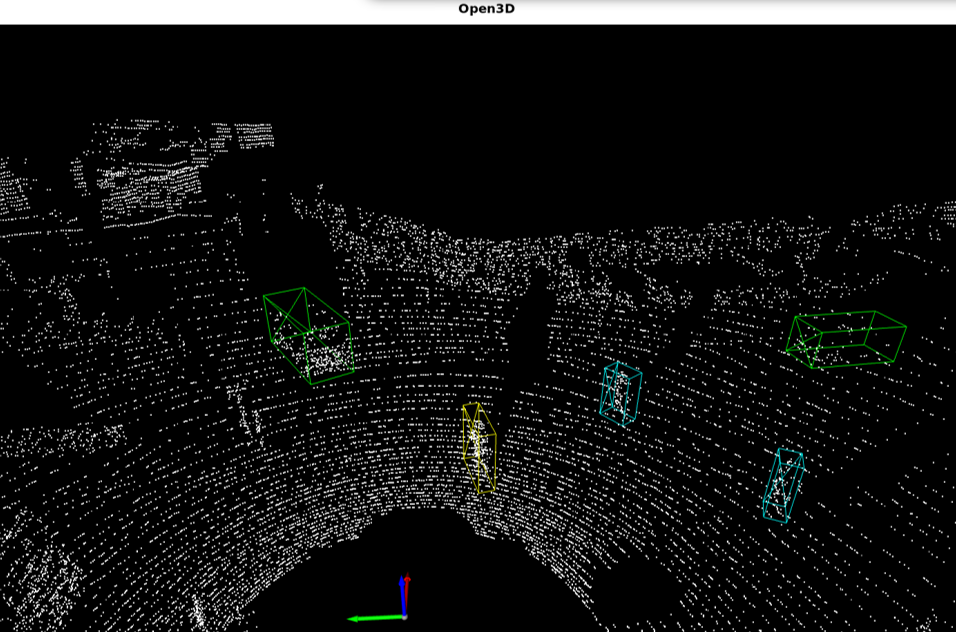}
    }
    \subfigure[Lane line detection results of the image using Gen-Lannet. The three sub-images are the original perspective, the aerial view, and the coordinate map.]{
        \includegraphics[width=.45\textwidth]{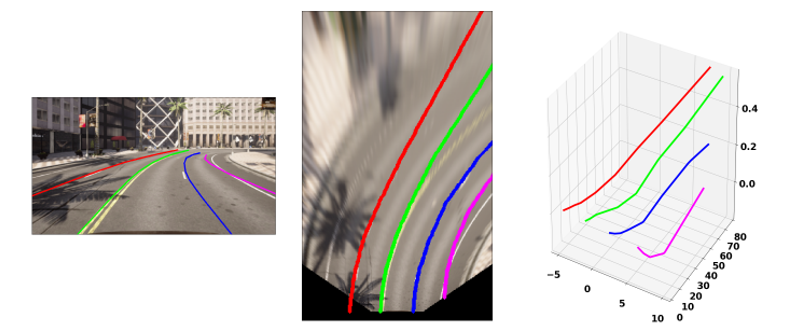}
    }
    \caption{Object detection approaches.}
\end{figure}

In the detection of pedestrians and vehicles, we use point-RCNN\cite{openpcdet2020} to process Point cloud data and obtain boundary information of each vehicle and pedestrian.

In the detection of lane lines, Gen-Lanenet\cite{LANE} was used to process the image data to obtain the spatial distribution of each lane.

\subsubsection{Object Tracking and Prediction}

In the tracking of pedestrians and vehicles, we use AB3DMOT\cite{AB3DMOT} to perform state estimation and data association for boundary information and obtain the object ID to which each boundary information belongs.

Next, we use a simple prediction method that uses exponential smoothing, in which the predicted motion is calculated and updated from the past trajectory.
\begin{equation}
    S_{n+1} = t * S_{n} + (1-t) * S_{n-1}
\end{equation}

\subsubsection{Data Fusion}

\begin{figure}[!ht]
    \centering
    \includegraphics[width=.3\textwidth]{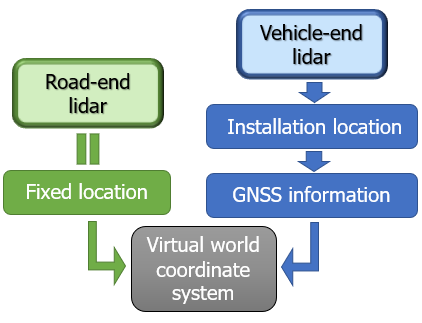}
    \caption{Unified scheme of coordinate system.}
\end{figure}

As for the data fusion method, we consider using a feature-level fusion scheme to share and combine the point cloud information sensed by lidar at the vehicle end and the road end.

To unify the coordinate system of the point cloud, the first thing to do is to establish a unified virtual coordinate system \cite{lan2016developmentUAV,lan2016developmentVR}. We choose the coordinate system of CARLA world as the virtual coordinate system. Then we change the point cloud coordinate system to the virtual coordinate system.

For the road-end lidar, it is assumed that we can directly know its position $L_{road}$ relative to the virtual coordinate system when installing it, then for each data $p_{road}$, it's final coordinate $P_{road}$ is

\begin{equation}
    P_{road} = p_{road} + L_{road}
\end{equation}

For vehicle-end lidar, we first need to align it to the vehicle coordinate system according to the installation position $D_{vehicle}$ of lidar relative to the vehicle. Then according to the real-time position $L_{vehicle}$ of the vehicle relative to the virtual coordinate system and the yaw angle $\theta$, the vehicle coordinate system is transformed into a unified virtual coordinate system. So for each data $p_{vehicle}$, its final coordinate $P_{vehicle}$ is

\begin{equation}
     P_{vehicle} = \mathbf{R} (p_{vehicle} + D_{vehicle}) + L_{vehicle} 
\end{equation}

where 
$$ \mathbf{R} = \begin{bmatrix}
 cos\theta & -sin\theta & 0\\
 sin\theta & cos\theta & 0\\
 0 & 0 & 1
\end{bmatrix} $$

After the unification of the lidar coordinate system at the opposite end of the vehicle and the road, the point cloud data of the two can be combined to participate in the subsequent object detection process as a whole.

\begin{figure*}[!ht]
    \centering
    \subfigure[Curve road test scenario. The blue circle is our vehicle, the green circle is the installation location of the road end equipment, and the red circle is the target vehicle.]{
        \includegraphics[width=.28\linewidth]{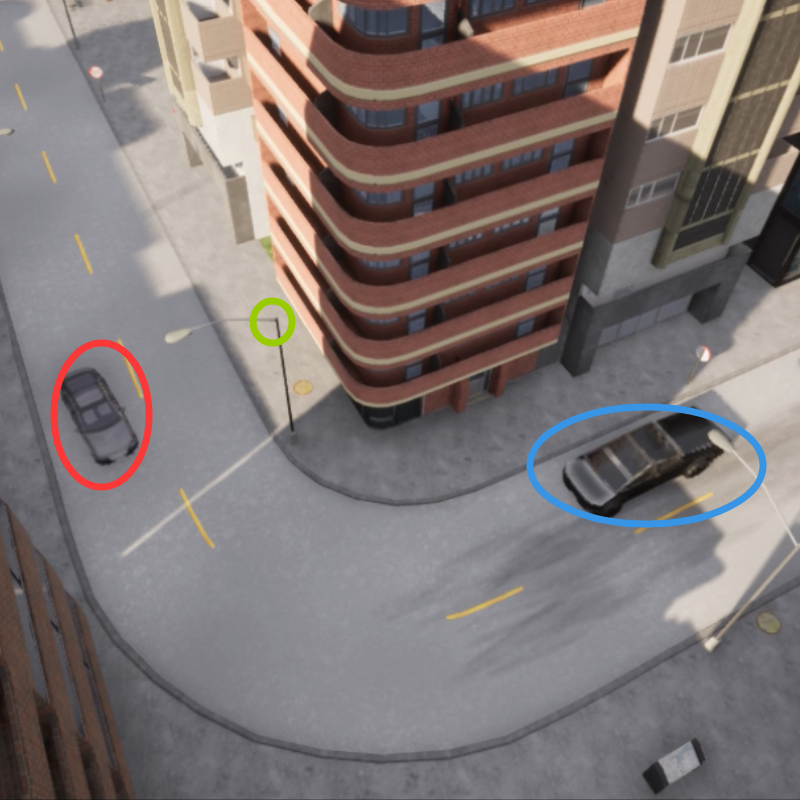}
        \label{test1:scene}
    }
    \subfigure[Blind area test scenario. The blue circle is our vehicle, the green circle is where the road end lidar is installed, and the red circle is a pedestrian.]{
        \includegraphics[width=.28\linewidth]{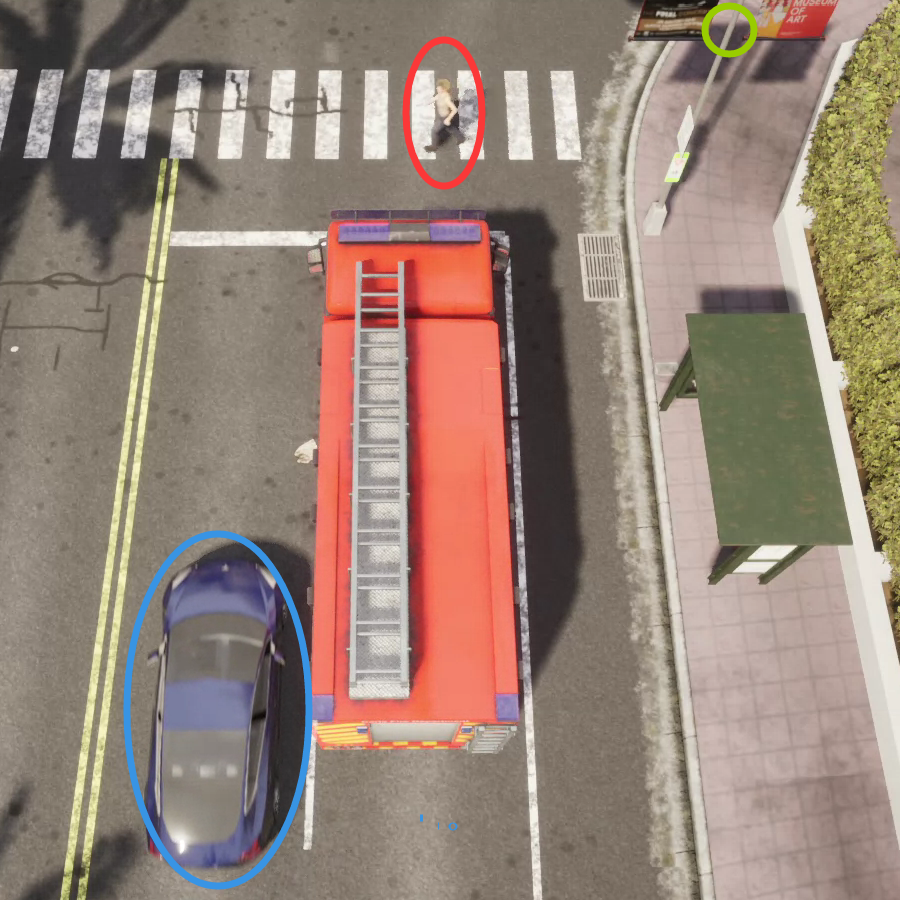}
        \label{test2:scene}
    }
    \subfigure[Blind area test scenario. The blue circle is our vehicle, the green circle is where the road end lidar is installed, and the red circle is a pedestrian.]{
        \includegraphics[width=.28\linewidth]{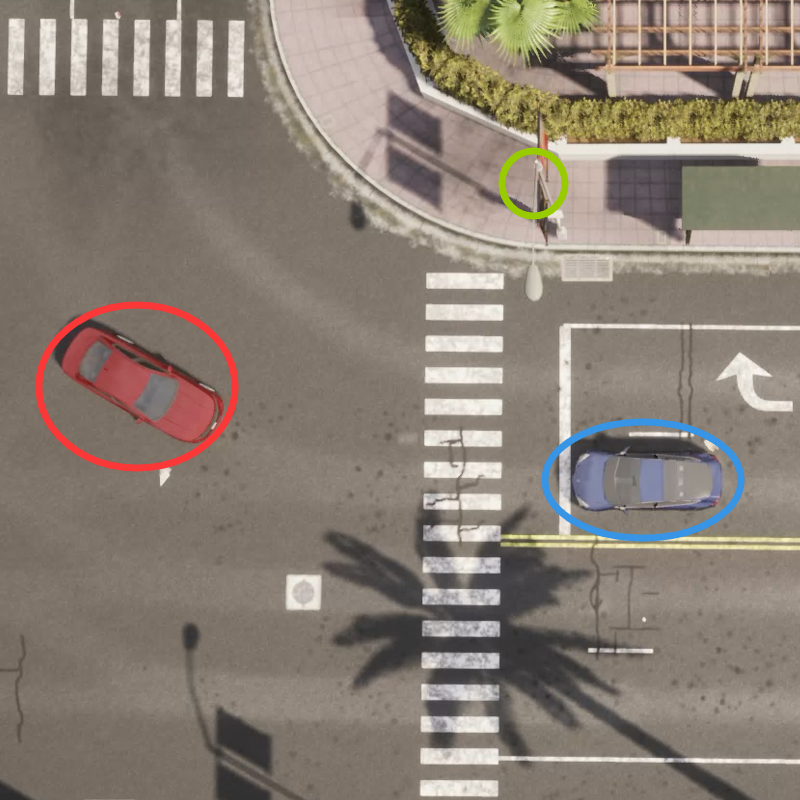}
        \label{test3:scene}
    }
    \caption{Three test scenarioes}
\end{figure*}
\section{Experimental setup}
\label{sec:setup}


We have designed three kinds of tests to verify our vehicle-road cooperative intelligent perception system, which will be tested for the perceptual ability of the system.

\subsection{Extended Range}

The autonomous driving of a single-vehicle can only perceive objects within a certain range due to the limited sensing range of sensors. However, in autonomous driving with vehicle-road cooperation, the vehicle can obtain the information perceived by the lidar at the far end of the road, so that the vehicle's perception range can break through the limitation of the sensor and reach farther.

In this test, we designed a vehicle driving on a curve road as shown in\ref{test1:scene}. Road end lidar is installed on the side of the curve to sense the traffic flow at both ends of the road. Our vehicles entered from one end of the curve, but the lidar of the vehicle could not directly perceive the vehicles at the other end of the curve due to the limitation of the sensing range.

We plan to use the vehicle-road cooperation perception system, in which the end of the road can share the perception information with our vehicles after perceiving the vehicles at the other end so that our vehicles can perceive the coming vehicles in the curves in advance.

\subsection{Supplemental Blind area}

Since the autonomous driving of a single-vehicle can only perceive the environment from one perspective, there will be a blind area due to the occlusion of other objects, which is easy to cause traffic accidents. However, for autonomous driving with vehicle-road cooperation, road-end devices and vehicles can perceive objects from different perspectives, thus reducing blind areas and increasing the safety of autonomous driving.

We chose a typical blind area scene for this test. In this scenario shown\ref{test2:scene}, when our vehicle is passing the intersection, a pedestrian is crossing the zebra crossing, but due to the obstruction of the sidecar body, the vehicle cannot sense the pedestrian. At the intersection, we set up the road end lidar, to perceive pedestrians and share the perception information with our vehicles.

For the single-vehicle autonomous driving system, this scenario is dangerous for traffic accidents, because the sidecar body obstruction is always unavoidable until it is about to collide with the pedestrian. However, for the vehicle-road cooperative system, the blind area problem is greatly reduced. Pedestrians crossing the road should be perceived and identified by the system to avoid collisions.

\subsection{Improved Accuracy}

the autonomous driving of a single-vehicle can usually only perceive the position of the target from one side, which can be inaccurate. However, for autonomous driving with vehicle-road cooperation, objects are viewed from multiple sides, which allows for more information and a more accurate estimate of position and pose.

In this test, we set the target vehicle at a position that both our vehicle lidar and the road-end lidar can perceive, as shown in \ref{test3:scene}. Both our vehicle and the road-end can obtain the estimation of the target vehicle pose respectively. Through the vehicle-road cooperative system, the information perceived by both vehicles can be shared, which we hope can improve the accuracy of target perception.

\section{Results}
\label{sec:results}


\subsection{Extended Range}

\begin{figure}[htbp]
    \centering
    \subfigure[The perception result. The blue point cloud is perceived by the road end, the green point cloud is perceived by the vehicle end, and the blue box is the boundary detected by the road end.]{
        \includegraphics[width=.2\textwidth]{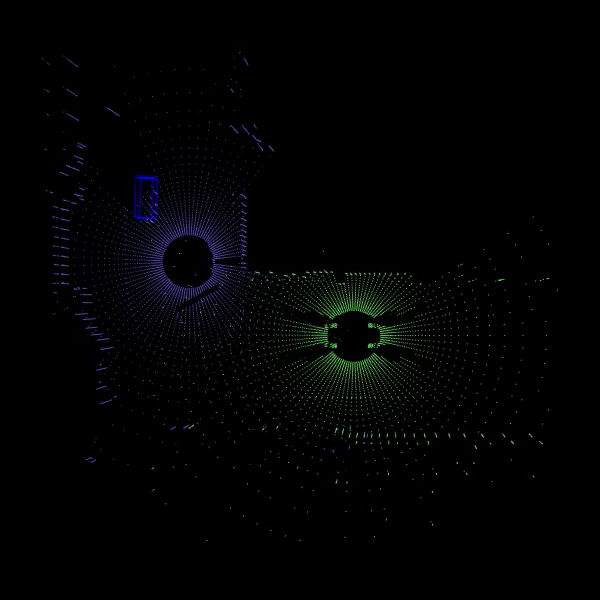}
        \label{test1:detection}
    }
    \subfigure[Schematic diagram of perceptual range. The vehicle's perception of 20 meters is expanded to 40 meters through the shared perception data at the road end, so the target vehicle can be sensed at 30 meters.]{
        \includegraphics[width=.25\textwidth]{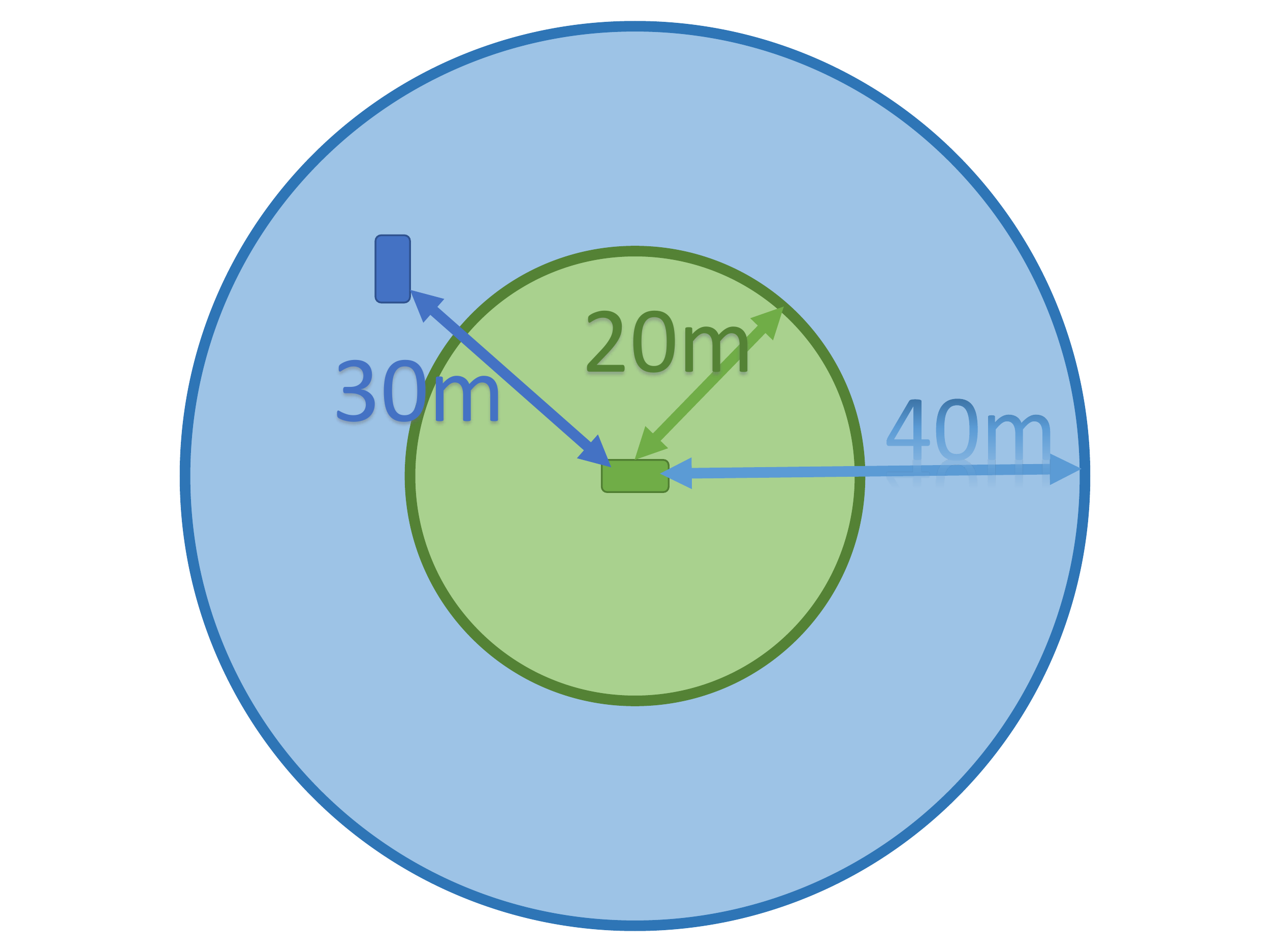}
        \label{test1:detection}
    }
    \caption{Test1}
\end{figure}

When the system runs in this scenario, we can see in \ref{test1:detection} that the target vehicle is sensed and recognized by the road-end equipment when it enters the curve. At this moment, in our vehicle's view, the target vehicle is still out of range of perception.

With the sharing of perceptual information by the equipment at the road end, our vehicle obtained the location information and boundary information of the target vehicle at the other end of the road. From the perspective of our vehicle, we can see that even though the sensing range of vehicle lidar is only 20 meters, the position and pose information of the target vehicle 30 meters away can be obtained in advance through the vehicle-road cooperation system.

In this scenario, through the vehicle's cooperative intelligent perception system, the information of the target vehicle is acquired by our vehicle outside the perception range of our vehicle, so that the perception range of our vehicle is expanded.

The expansion of perception range can solve the problem of traffic jams to some extent because vehicles can sense the traffic flow on the road ahead in advance and make more efficient strategies such as changing lanes accordingly.

\subsection{Supplemental Blind area}

\begin{figure}[!ht]
    \centering
    \subfigure[The perception result. The blue point cloud is perceived by the road end, and the green point cloud is perceived by the vehicle end.]{
        \includegraphics[width=.3\textwidth]{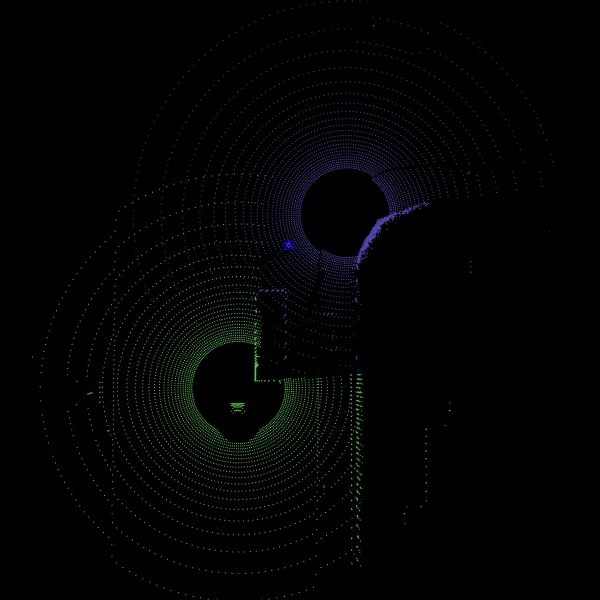}
        \label{test2:result}
    }
    \subfigure[Distance between pedestrians and vehicles changes over time. The green and blue zones represent pedestrians sensed at the road end and the vehicle end, respectively, and the orange zone represents that the collision occurred.]{
        \includegraphics[width=.45\textwidth]{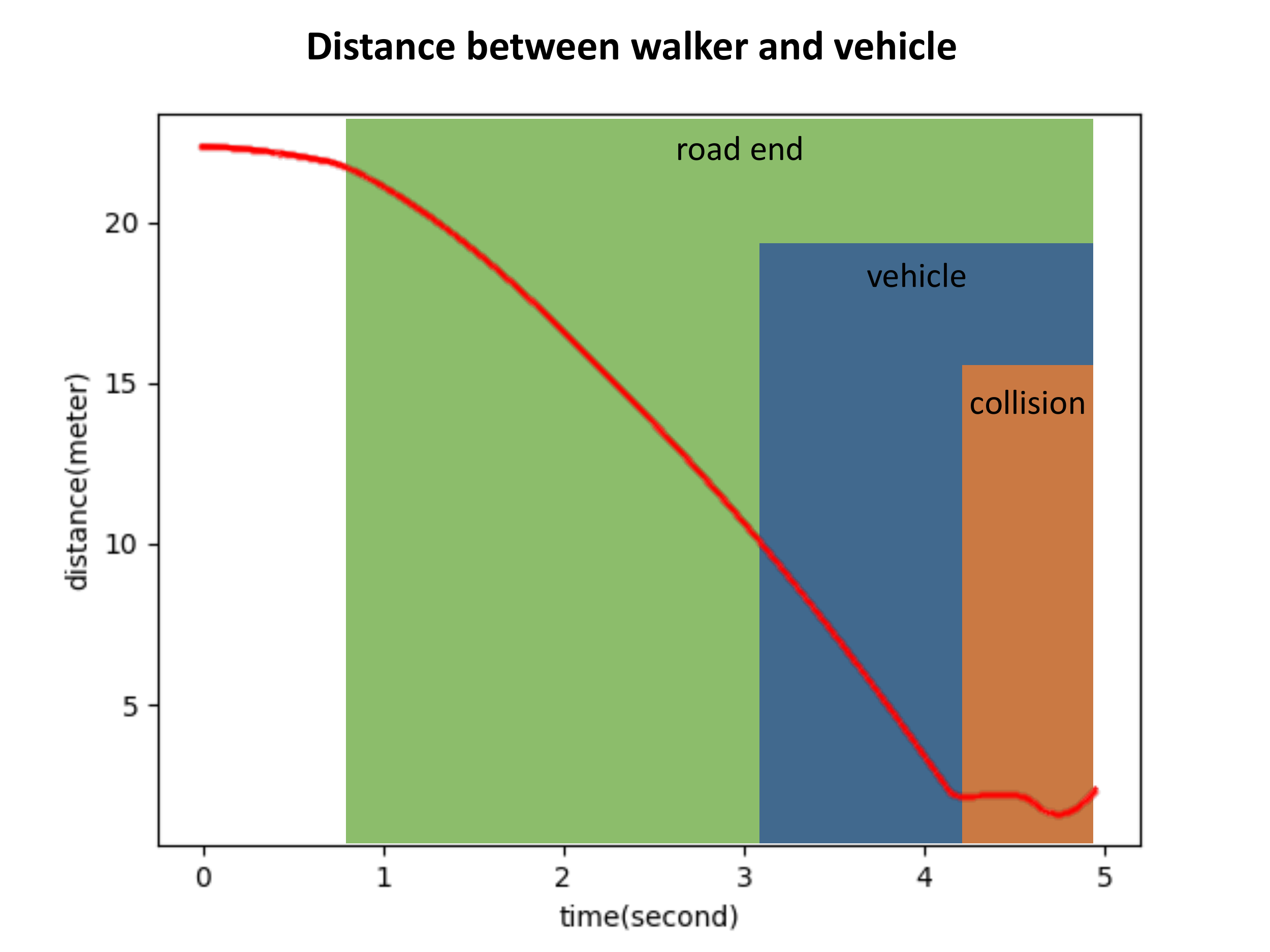}
        \label{test2:graph}
    }
    \caption{Test2}
\end{figure}

When the system runs in this scenario, we can get the perception results in \ref{test2:result}. It can be seen from the results that the original vehicle lidar point cloud could not perceive pedestrians because most of the perception range of the right front of the vehicle was blocked by the sidecar body. The view from 10 degrees to 160 degrees on the right side of the vehicle is blocked, while pedestrians appear 30 degrees on the right side of the vehicle. As can be seen from \ref{test2:graph}, the braking distance and reaction time of the vehicle from perceiving a pedestrian to the collision is only 8 meters and 1 second, which is insufficient for the 12.7 meters braking distance required by the vehicle traveling at 30 km/h.

However, with the help of the vehicle-road cooperation system, the perception information at the road end is shared with the vehicle end when the vehicle passes the intersection, which is combined with the vehicle end through data fusion to supplement the perception information of the covered area so that the vehicle end can indirectly perceive the pedestrian and avoid accidents before 20 meters.

This test demonstrates the superiority of the vehicle-road cooperative system in terms of safety. In the case of only relying on the vehicle's sensor, the vehicle can not perceive the covered target, resulting in a potential safety hazard. However, with the supplement of the vehicle-road cooperation system, the vehicle perception information is broadened, the blind area of the visual field is reduced, and the hidden danger is eliminated.

In this scenario, through the vehicle-road cooperative intelligent perception system, the blind area of the perceptual information of the vehicle is supplemented, which solves the thorny blind area problem of single-vehicle autonomous driving.

\subsection{Improved Accuracy}

\begin{figure}[htbp]
    \centering
    \subfigure[The perception result. The blue point cloud is perceived by the road end, and the green point cloud is perceived by the vehicle end.]{
        \includegraphics[width=.22\textwidth]{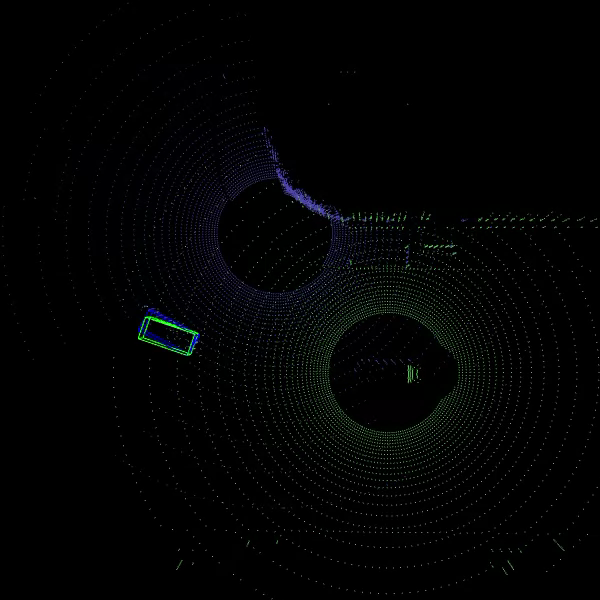}
        \label{test3:result}
    }
    \subfigure[Comparison of detection accuracy. Red is the ground truth, blue is the vehicle detection result, green is the road end detection result, and orange is the fusion result.]{
        \includegraphics[width=.23\textwidth]{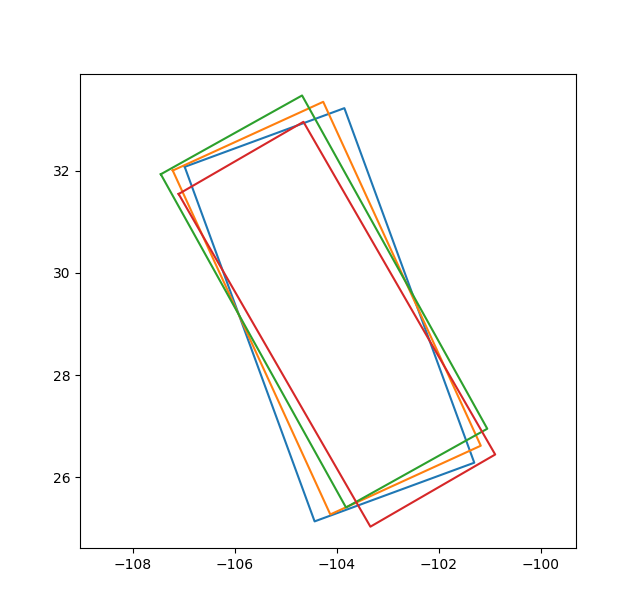}
        \label{test3:mat}
    }
    \caption{Test3}
\end{figure}

When the system runs in this scenario, we can get the perception results in \ref{test3:result}. It can be seen that the vehicle and the road end get different boundaries to the target vehicle through perception.

We consider the boundary overlap as our evaluation indicator. Boundary overlap refers to the ratio of the intersection area between the detected boundary and the real boundary of the target to the real boundary area. After calculation, the boundary overlap of the vehicle is 92.9\%, that of the road end is 92.8\%, and that of the fusion result is 93.7\%.

In this scenario, through the vehicle-road cooperative intelligent perception system, the boundary overlap perceived by vehicles to target vehicles increased from 92.9\% to 93.7\%, indicating that vehicle-road cooperation can improve the accuracy of perception. 

With higher perceptual accuracy, autonomous vehicles can perform more precise operations, increasing the safety and efficiency of autonomous driving.

\section{Discussion}
\label{sec:discussion}


In this project, we designed and implemented the vehicle-road cooperation intelligent perception system, including its processing flow and data fusion scheme. After that, we designed the feasibility and safety test scenarios for the vehicle-road cooperative system. Then, we used the test scenes we designed to test our vehicle-road cooperative intelligent perception system. The test results not only show that our system can realize the intelligent perception of vehicles and pedestrians, but also prove the feasibility and safety of the system by expanding the perception range, supplementing the perception blind area and improving perception accuracy. This also proves the superiority of vehicle-road collaboration over single-vehicle autonomous driving.

This project can be improved as follows: first, the system can adopt multi-mode sensor information and realize a multi-mode fusion scheme at the same time, to achieve a better perception effect. Second, more test scenarios can be designed later to increase the complexity of the system so that it can cope with more complex situations.











\bibliographystyle{IEEEtran}
\bibliography{bibliography}

\begin{thebibliography}{10}
\providecommand{\url}[1]{#1}
\csname url@samestyle\endcsname
\providecommand{\newblock}{\relax}
\providecommand{\bibinfo}[2]{#2}
\providecommand{\BIBentrySTDinterwordspacing}{\spaceskip=0pt\relax}
\providecommand{\BIBentryALTinterwordstretchfactor}{4}
\providecommand{\BIBentryALTinterwordspacing}{\spaceskip=\fontdimen2\font plus
\BIBentryALTinterwordstretchfactor\fontdimen3\font minus
  \fontdimen4\font\relax}
\providecommand{\BIBforeignlanguage}[2]{{%
\expandafter\ifx\csname l@#1\endcsname\relax
\typeout{** WARNING: IEEEtran.bst: No hyphenation pattern has been}%
\typeout{** loaded for the language `#1'. Using the pattern for}%
\typeout{** the default language instead.}%
\else
\language=\csname l@#1\endcsname
\fi
#2}}
\providecommand{\BIBdecl}{\relax}
\BIBdecl

\bibitem{WangLonghe2019Vcci}
L.~Wang, B.~Ai, D.~He, K.~Guan, J.~Zhang, J.~Kim, and Z.~Zhong,
  ``\BIBforeignlanguage{eng}{Vehicle-to-infrastructure channel characterization
  in urban environment at 28 ghz},'' \emph{\BIBforeignlanguage{eng}{China
  communications}}, vol.~16, no.~2, pp. 36--48, 2019.

\bibitem{V2I}
H.~Kaur and Meenakshi, ``Analysis of vanet geographic routing protocols on real
  city map,'' in \emph{2017 2nd IEEE International Conference on Recent Trends
  in Electronics, Information Communication Technology (RTEICT)}, 2017, pp.
  895--899.

\bibitem{lan2022semantic}
G.~Lan, T.~Liu, X.~Wang, X.~Pan, and Z.~Huang, ``A semantic web technology
  index,'' \emph{Scientific Reports}, vol.~12, no.~1, pp. 1--10, 2022.

\bibitem{lan2022class}
G.~Lan, Z.~Gao, L.~Tong, and T.~Liu, ``Class binarization to neuroevolution for
  multiclass classification,'' \emph{Neural Computing and Applications}, pp.
  1--18, 2022.

\bibitem{V2P}
N.~Liu, M.~Liu, J.~Cao, G.~Chen, and W.~Lou, ``When transportation meets
  communication: V2p over vanets,'' in \emph{2010 IEEE 30th International
  Conference on Distributed Computing Systems}, 2010, pp. 567--576.

\bibitem{lan2018directed}
G.~Lan, M.~Jelisavcic, D.~M. Roijers, E.~Haasdijk, and A.~E. Eiben, ``Directed
  locomotion for modular robots with evolvable morphologies,'' in
  \emph{International Conference on Parallel Problem Solving from
  Nature}.\hskip 1em plus 0.5em minus 0.4em\relax Springer, 2018, pp. 476--487.

\bibitem{V2X}
P.~Panchapakesan, ``\BIBforeignlanguage{English}{Overview of lte based cellular
  v2x communication},'' \emph{\BIBforeignlanguage{English}{Telecom Business
  Review}}, vol.~10, no.~1, pp. 23--30, 2017, copyright Publishing India Group
  2017.

\bibitem{xu2019online}
H.~Xu, G.~Lan, S.~Wu, and Q.~Hao, ``Online intelligent calibration of cameras
  and lidars for autonomous driving systems,'' in \emph{2019 IEEE Intelligent
  Transportation Systems Conference (ITSC)}.\hskip 1em plus 0.5em minus
  0.4em\relax IEEE, 2019, pp. 3913--3920.

\bibitem{wu2022sparse}
X.~Wu, L.~Peng, H.~Yang, L.~Xie, C.~Huang, C.~Deng, H.~Liu, and D.~Cai,
  ``Sparse fuse dense: Towards high quality 3d detection with depth
  completion,'' \emph{arXiv preprint arXiv:2203.09780}, 2022.

\bibitem{wu2021tracklet}
H.~Wu, Q.~Li, C.~Wen, X.~Li, X.~Fan, and C.~Wang, ``Tracklet proposal network
  for multi-object tracking on point clouds,'' in \emph{Proceedings of the
  International Joint Conference on Artificial Intelligence (IJCAI)}, 2021, pp.
  1165--1171.

\bibitem{lan2019evolutionary}
G.~Lan, J.~Chen, and {Eiben, A. E.}, ``Evolutionary predator-prey robot
  systems: From simulation to real world,'' in \emph{Proceedings of the Genetic
  and Evolutionary Computation Conference Companion}, 2019, pp. 123--124.

\bibitem{lan2019simulated}
G.~Lan, J.~Chen, and A.~E. Eiben, ``Simulated and real-world evolution of
  predator robots,'' in \emph{2019 IEEE Symposium Series on Computational
  Intelligence (SSCI)}.\hskip 1em plus 0.5em minus 0.4em\relax IEEE, 2019, pp.
  1974--1981.

\bibitem{s19091967}
Z.~Xiao, D.~Yang, F.~Wen, and K.~Jiang, ``A unified multiple-target positioning
  framework for intelligent connected vehicles,'' \emph{Sensors}, vol.~19,
  no.~9, 2019.

\bibitem{8943388}
Z.~Wang, Y.~Wu, and Q.~Niu, ``Multi-sensor fusion in automated driving: A
  survey,'' \emph{IEEE Access}, vol.~8, pp. 2847--2868, 2020.

\bibitem{lan2017development}
G.~Lan, J.~Liang, G.~Liu, and Q.~Hao, ``Development of a smart floor for target
  localization with bayesian binary sensing,'' in \emph{2017 IEEE 31st
  International Conference on Advanced Information Networking and Applications
  (AINA)}.\hskip 1em plus 0.5em minus 0.4em\relax IEEE, 2017, pp. 447--453.

\bibitem{lan2016developmentVR}
G.~Lan, Z.~Luo, and Q.~Hao, ``Development of a virtual reality teleconference
  system using distributed depth sensors,'' in \emph{2016 2nd IEEE
  International Conference on Computer and Communications (ICCC)}.\hskip 1em
  plus 0.5em minus 0.4em\relax IEEE, 2016, pp. 975--978.

\bibitem{AITEST}
H.~J. Vishnukumar, B.~Butting, C.~Müller, and E.~Sax, ``Machine learning and
  deep neural network — artificial intelligence core for lab and real-world
  test and validation for adas and autonomous vehicles: Ai for efficient and
  quality test and validation,'' in \emph{2017 Intelligent Systems Conference
  (IntelliSys)}, 2017, pp. 714--721.

\bibitem{lan2019evolving}
G.~Lan, L.~De~Vries, and S.~Wang, ``Evolving efficient deep neural networks for
  real-time object recognition,'' in \emph{2019 IEEE Symposium Series on
  Computational Intelligence (SSCI)}.\hskip 1em plus 0.5em minus 0.4em\relax
  IEEE, 2019, pp. 2571--2578.

\bibitem{lan2018real}
G.~Lan, J.~Benito-Picazo, D.~M. Roijers, E.~Dom{\'\i}nguez, and A.~E. Eiben,
  ``Real-time robot vision on low-performance computing hardware,'' in
  \emph{2018 15th international conference on control, automation, robotics and
  vision (ICARCV)}.\hskip 1em plus 0.5em minus 0.4em\relax IEEE, 2018, pp.
  1959--1965.

\bibitem{lan2022time}
G.~Lan, J.~M. Tomczak, D.~M. Roijers, and A.~E. Eiben, ``Time efficiency in
  optimization with a bayesian-evolutionary algorithm,'' \emph{Swarm and
  Evolutionary Computation}, vol.~69, p. 100970, 2022.

\bibitem{lan2021learning2}
G.~Lan, M.~De~Carlo, F.~van Diggelen, J.~M. Tomczak, D.~M. Roijers, and A.~E.
  Eiben, ``Learning directed locomotion in modular robots with evolvable
  morphologies,'' \emph{Applied Soft Computing}, vol. 111, p. 107688, 2021.

\bibitem{lan2021learning1}
G.~Lan, M.~van Hooft, M.~De~Carlo, J.~M. Tomczak, and A.~E. Eiben, ``Learning
  locomotion skills in evolvable robots,'' \emph{Neurocomputing}, vol. 452, pp.
  294--306, 2021.

\bibitem{CARLA}
A.~Dosovitskiy, G.~Ros, F.~Codevilla, A.~Lopez, and V.~Koltun, ``{CARLA}: {An}
  open urban driving simulator,'' in \emph{Proceedings of the 1st Annual
  Conference on Robot Learning}, 2017, pp. 1--16.

\bibitem{openpcdet2020}
O.~D. Team, ``Openpcdet: An open-source toolbox for 3d object detection from
  point clouds,'' \url{https://github.com/open-mmlab/OpenPCDet}, 2020.

\bibitem{LANE}
Y.~Guo, G.~Chen, P.~Zhao, W.~Zhang, J.~Miao, J.~Wang, and T.~E. Choe,
  ``Gen-lanenet: A generalized and scalable approach for 3d lane detection,''
  2020.

\bibitem{AB3DMOT}
X.~Weng, J.~Wang, D.~Held, and K.~Kitani, ``{3D Multi-Object Tracking: A
  Baseline and New Evaluation Metrics},'' \emph{IROS}, 2020.

\bibitem{lan2016developmentUAV}
G.~Lan, J.~Sun, C.~Li, Z.~Ou, Z.~Luo, J.~Liang, and Q.~Hao, ``Development of
  uav based virtual reality systems,'' in \emph{2016 IEEE International
  Conference on Multisensor Fusion and Integration for Intelligent Systems
  (MFI)}.\hskip 1em plus 0.5em minus 0.4em\relax IEEE, 2016, pp. 481--486.

\end{thebibliography}

\end{document}